\documentclass[runningheads]{llncs}
\usepackage{graphicx}
\usepackage{amsmath}
\usepackage{amsfonts}
\usepackage{multirow}
\usepackage{algorithm}
\usepackage{cancel}
\usepackage{arydshln}
\usepackage[noend]{algpseudocode}
\usepackage[caption=false]{subfig}
\newcommand{\donotdisplay}[1]{}
\newlength{\lgCase}
\graphicspath{{Figures/}}

\usepackage{url}

\begin{document}
%
\title{Dirichlet-Survival Process: Scalable Inference of Topic-Dependent Diffusion Networks}
\titlerunning{Dirichlet-Survival Process}
%
\author{Gaël Poux-Médard\inst{1}\orcidID{0000-0002-0103-8778} \and
Julien Velcin\inst{1}\orcidID{0000-0002-2262-045X} \and
Sabine Loudcher\inst{1}\orcidID{0000-0002-0494-0169}}
\authorrunning{G. Poux-Médard et al.}
%
\institute{$^1$ Université de Lyon, Lyon 2, ERIC UR 3083, 5 avenue Pierre Mendès France, F69676 Bron Cedex, France\\
\email{gael.poux-medard@univ-lyon2.fr}\\
\email{julien.velcin@univ-lyon2.fr}\\
\email{sabine.loudcher@univ-lyon2.fr}\\
}
\maketitle              
\begin{abstract}
Information spread on networks can be efficiently modeled by considering three features: documents' content, time of publication relative to other publications, and position of the spreader in the network. Most previous works model up to two of those jointly, or rely on heavily parametric approaches. Building on recent Dirichlet-Point processes literature, we introduce the Houston (Hidden Online User-Topic Network) model, that jointly considers all those features in a non-parametric unsupervised framework. It infers dynamic topic-dependent underlying diffusion networks in a continuous-time setting along with said topics. It is unsupervised; it considers an unlabeled stream of triplets shaped as \textit{(time of publication, information's content, spreading entity)} as input data. Online inference is conducted using a sequential Monte-Carlo algorithm that scales linearly with the size of the dataset. Our approach yields consequent improvements over existing baselines on both cluster recovery and subnetworks inference tasks.

\end{abstract}

\section{Introduction}
\subsection{Overview of the contribution}
Over the last decades, information spread patterns have become more and more complicated. The volume of data that flows on social networks keeps increasing every day that passes, and results in complex diffusion processes that can be described by many factors. However, recent advances suggest that documents complex diffusion processes can be efficiently modeled considering only three variables: their publication date (when), the publisher (who) and their semantic content (what).
The idea of considering these three factors is not novel. However, most of the models that tackle diffusion problems tend to consider up to two of these, but seldom the three parameters. 

We introduce the Houston model, that tackles the problem by jointly inferring clusters of textual documents spreading online \textit{and} the subnetworks they spread on. Our method builds on recent Dirichlet-Point processes advances \cite{Du2015DHP,Valera2017HDHP,Poux2021PDHP,Poux2023MPDHP}. To the best of our knowledge, it is the first model that considers semantic content, publication dynamics and the network of spreading documents in an online, non-parametric and unsupervised way.

\begin{figure}
    \centering
    \includegraphics[width=\columnwidth]{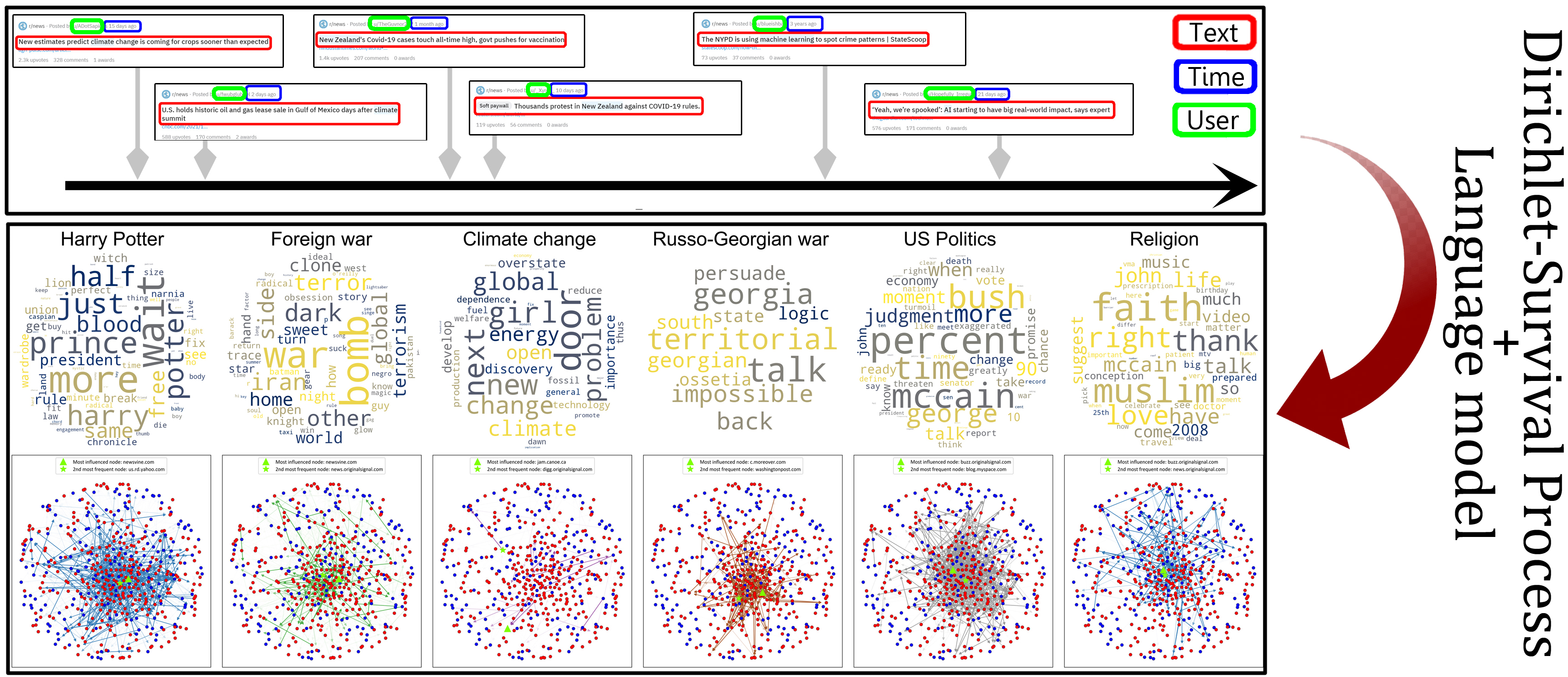}
    \caption{From a stream of textual documents, we model the underlying topic-dependent diffusion subnetworks. Inference is unsupervised, non-parametric and conducted online, meaning data is processed sequentially. Results in the bottom row come from the application of our method to the Memetracker dataset \cite{Leskovec2009Memetracker}. Nodes colors represent traditional medias (red) and blog (blue).}
    \label{fig:task-houston}
\end{figure}

\subsection{Related works}
It has been underlined on several occasions that efficiently modeling information diffusion involves accounting for the network's structure \cite{Larremore2012StatisticalPropertiesAvalanches,Poux2020InfluentialSpreaders}, publication times \cite{Du2012KernelCascade,GomezRodriguez2011NetRate} and documents' content \cite{Du2013TopicCascade,He2015HawkesTopic}.
Some approaches consider sequentially all three factors. Typically, they first infer topics based on documents content, and only then they use this information to infer the latent diffusion subnetworks \cite{Choudhari2018TopicalInteractionHMHP,Du2013TopicCascade,He2015HawkesTopic,Suny2018InferringMD,Wang2012Monet,Shuang2013}. 
The work the closest to ours \cite{Barbieri2017SurvivalFactorization} is, to our knowledge, the only one that jointly models documents' content, dynamics and structure. It develops an unsupervised topic-dependent network inference method. The approach breaks down the topic-aware diffusion into two factors: each node is assumed to have a given sensitivity to a topic, and a certain authority on them. Given this assumption, the authors develop a parametric prior on the probability for a diffusion cascade to belong to a given topic. The textual content (or side information) is then accounted for using a homogeneous Poisson textual model \cite{Mei2007PoissonLang}, combined with the above prior. The model is optimized using an EM algorithm.
However, the optimization algorithm is not designed for online optimization --data cannot be added sequentially--, and topics optimization is parametric --the number of topics must be provided.

\section{Model}
\subsection{Background}
To answer these limitations, we build a Dirichlet-Survival process that can be used as a non-parametric Bayesian prior for online inference. The Dirichlet-Survival prior is created by merging Dirichlet processes with Point processes. The method has been explored by combining Hawkes processes to several variants of Dirichlet processes (hierarchical \cite{Valera2017HDHP}, mixed membership \cite{Tan2018IBHP}, powered \cite{Poux2021PDHP}, multivariate \cite{Poux2023MPDHP}). However, no work considered the combination with other point processes than the Hawkes process. Our approach using Survival analysis explores this new connection; it allows us to design an optimization algorithm (Sequential Monte Carlo) for online non-parametric topics-aware diffusion subnetworks inference (the number of topics/subnetworks does not have to be chosen in advance).

In \cite{GomezRodriguez2013SurvivalAnalysis}, the authors show that a large part of the literature on underlying diffusion network inference \cite{Du2012KernelCascade,GomezRodriguez2011NetRate,GomezRodriguez2013InfoPath,GomezRodriguez2013SurvivalAnalysis,Nickel2021LazyHawkes,Wang2012Monet} can be expressed as special cases of a counting point process. The method allows to infer dynamic underlying diffusion networks using convex optimization tools.

\subsection{Dirichlet process and Survival analysis}

\subsubsection{Dirichlet process}
The Dirichlet process is used as a non-parametric prior distribution over clusters in many clustering algorithms. It can be written as follows:
\begin{equation}
    \label{eq-DP}
    P(s_i=k \vert \{s_m\}_{m=1, ..., n-1},\alpha_0) = \begin{cases} 
    \frac{N_k}{\alpha_0 + \sum_k^K N_k} \text{ if k = 1, ..., K}\\
    \frac{\alpha_0}{\alpha_0 + \sum_k^K N_k} \text{ if k = K+1}
    \end{cases}
\end{equation}
where $s_i$ is a variable that represents the cluster of the $i^{th}$ observation, $N_k = \vert \{ s_i \vert s_i=k \}_{i=1,...,n-1} \vert$ the population of cluster $k$, $K$ the total number of non-empty clusters and $\alpha_0$ a concentration hyper-parameter. The choice of $K+1$ means a new cluster is opened and $K$ in increased by 1. Note that references \cite{Poux2021PDHP,Poux2023MPDHP} use the powered version of this process \cite{Poux2021PDP}.

\subsubsection{Network inference model}
The edges of topic-dependent networks are inferred using the NetRate model \cite{GomezRodriguez2011NetRate}, which is part of a broad literature on underlying spreading networks inference \cite{Du2013TopicCascade,GomezRodriguez2011NetRate,GomezRodriguez2013InfoPath,GomezRodriguez2013SurvivalAnalysis,Wang2012Monet}. In particular in \cite{GomezRodriguez2013SurvivalAnalysis}, the authors demonstrate that all these models can be expressed as special cases of a counting point process. These processes take a collection of independent timestamped diffusion cascades $\Vec{c} = \{ (u_i^c, t_i^c) \}_i$ as input, where $u_i^c$ is the node on which the $i^{th}$ event occurred and $t_i^c$ the time at which it happened in cascade $c$. The process is entirely characterized by a hazard function $H(t_i^c \vert t_j^c, \alpha_{u_j^c,u_i^c})$, which is the instantaneous infection rate of $u_i^c$ at time $t_i^c$ by $u_j^c$ previously infected at time $t_j^c$, given it infection did not happen before $t_i$. In this paper, we express the hazard function as a constant ${H(t \vert t_i, \alpha) = \alpha}$, implying by definition that the probability of an event \textit{non} happening before a time $t$ given $t_i$ decays exponentially as $e^{-\alpha (t-t_i)}$. The associated convex likelihood of $\alpha$ can be found in \cite{GomezRodriguez2011NetRate} (Eq.7).

\subsection{Dirichlet-Survival process}
In \cite{Du2015DHP} the authors define the Dirichlet-Hawkes process by replacing the integer counts in Eq.\ref{eq-DP} by the intensity of a Hawkes process. It can be interpreted as replacing integers counts in Dirichlet Processes by non-integer time-dependent counts, encoded by the intensity of the point process. Here, we consider the hazard rate of the NetRate model instead to account for networks structure. Each node is associated to its own temporal point process, and counts are replaced by the number of times any neighbour has been infected, weighted according to time and to edges strength. Using the methodology introduced in \cite{Du2015DHP} and substituting the Hawkes process by the hazard rate of a survival model \cite{GomezRodriguez2013SurvivalAnalysis}, we make a yet unexplored bridge between Dirichlet processes and Survival analysis. 
We remind that \cite{GomezRodriguez2013SurvivalAnalysis} reformulates the work of \cite{Du2012KernelCascade,GomezRodriguez2011NetRate,GomezRodriguez2013InfoPath,Wang2012Monet} in terms of Survival analysis and associated counting processes; we settled on using NetRate here, but any of these models would fit as well in our approach. The point process nature of survival analysis discussed in \cite{GomezRodriguez2013SurvivalAnalysis} makes this extension sound with respect to previous works on Dirichlet-Point processes \cite{Du2015DHP,Valera2017HDHP,Poux2021PDHP,Tan2018IBHP}. 

Let $\mathbf{A}^{(k)}$ be the adjacency matrix of the subnetwork associated to cluster $k$, whose entries are $\alpha_{i,j}^{(k)}$. We define $(u_j^c, t_j^c)^{(k)}$ as an event of cascade $c$ observed on node $u_j$ at $t_j$ attributed to subnetwork $A^{(k)}$. We write the history of events in cascade $c$ attributed to the subnetwork $k$ as $\mathcal{H}_{i,c}^{(k)} = \{ (u_j^c, t_j^c)^{(k)} \}_{j: t_j<t_i}$. We note $\mathbf{\mathcal{H}_{i,c}} = \{ \mathcal{H}_{i,c}^{(k)} \}_k$ and $\mathbf{A} = \{ \mathbf{A}^{(k)} \}_k$.
We consider a new event from cascade $c$ observed on node $u_i^c$ at time $t_i^c$. At this point, the new event is not yet associated to any subnetwork. We write the Dirichlet-Survival prior probability for the new event to belong to subnetwork $k$:
\begin{equation}
    \label{eq-DSP}
        P(s_i=k \vert \mathbf{\mathcal{H}_{i,c}}, \mathbf{A}, \mathbf{\lambda_0}) =
        \begin{cases} 
        \frac{\lambda_0^{(k)} + \sum_{\mathcal{H}_{i,c}^{(k)}} H(t_i^c \vert t_j^c, \alpha_{u_j,u_i}^{(k)})}{\lambda_0^{(K+1)} + \sum_k^K \lambda_0^{(k)} + \sum_{\mathcal{H}_{i,c}^{(k)}} H(t_i^c \vert t_j^c, \alpha_{u_j,u_i}^{(k)})} \text{ if k = 1, ..., K}\\
        \frac{\lambda_0^{(K+1)}}{\lambda_0^{(K+1)} + \sum_k^K \lambda_0^{(k)} + \sum_{\mathcal{H}_{i,c}^{(k)}} H(t_i^c \vert t_j^c, \alpha_{u_j,u_i}^{(k)})} \text{ if k = K+1}
        \end{cases}
\end{equation}
We introduced a new parameter $\mathbf{\lambda_0} = \{ \lambda_0^{(k)} \}_{k=1,...,K+1}$, which translates the probability for a new observation not to have been triggered by any neighbour. It represents the probability that an event of cluster $k$ is exogenous \cite{He2015HawkesTopic,Myers2012ExternalInfluence}.

The Dirichlet-Survival prior is coupled to a sequential language model. For simplicity, we consider the bag-of-words Dirichlet-Multinomial model, as in \cite{Du2015DHP,Valera2017HDHP,Poux2021PDHP}; note that more refined sequential language models are also fit to our approach (Dynamic Topic Model \cite{Blei2006DynamicTopicModel}, online LDA \cite{AlSumait2008OnlineLDA}, online PLSA \cite{Bassiou2014OnlinePLSA}, etc.).

The input data is a stream of events. Each event takes the form of a triplet $(u_i^c,t_{i}^c,v_{i}^c)$, where $c$ is the cascade an event has been observed in, $u_i^c$ is the node corresponding to the event, $t_{i}^c$ is its publication time, and $v_{i}^c$ represents its textual content (e.G. words in a tweet or in a news article).
By combining the Dirichlet-Survival prior to the textual likelihood, we get the posterior distribution of the $i^{th}$ observation belonging to cluster (or subnetwork) $k$ as:
\begin{equation}
    \label{eq-posterior}
        P(s_i \vert v_{i}^c, \mathbf{N}, \mathbf{\mathcal{H}_{i,c}}, \mathbf{A}, \theta_0,\mathbf{\lambda_0})
        \propto \underbrace{P(v_{i}^c \vert s_i, \mathbf{N}, \theta_0)}_{\text{Dirichlet-Multinomial}} \times \underbrace{P(s_i \vert \mathbf{\mathcal{H}_{i,c}}, \mathbf{A},\mathbf{\lambda_0})}_{\text{Dirichlet-Survival prior (Eq.\ref{eq-DSP}})}
\end{equation}
where $\vec{N}$ contains the words counts within each cluster, $v_{i}^c$ contains the words count in document $i$, and $\theta_0$ the concentration parameter of the model.

Finally, inference is conducted using a Sequential Monte Carlo algorithm similar to \cite{Du2015DHP,Valera2017HDHP,Poux2021PDHP}. We perform several parallel runs on the same data stream. Within each run, each new observation in the stream is assigned to a cluster according to Eq.~\ref{eq-posterior}. The adjacency matrix $\mathbf{A}$ is then updated by optimizing the convex likelihood associated to the NetRate point process (Eq.~7 in  \cite{GomezRodriguez2011NetRate}). Finally, we compute the likelihood of the language model for each run; runs that have a likelihood lesser than a threshold are discarded and replaced by more likely ones. The process is repeated until the end of the data stream. According to this algorithm, Eq.~\ref{eq-DP}, and introducing a cutoff on the exponential hazard function (observations older than a time $t_{old}$ are ignored), the optimization runs in $\mathcal{O}(N_{obs}N_{runs}(N_{nodes}+K))$ where $N_{part}$ is the number of particles, $N_{nodes}$ is the maximum network size and $K$ the number of clusters (typically $N_{runs} \ll K \ll N_{nodes}$). Inference hence scales linearly with the size of the dataset.

We point out that the Dirichlet-Survival process is not about refining complex diffusion models such as \cite{Barbieri2017SurvivalFactorization,Choudhari2018TopicalInteractionHMHP,Suny2018InferringMD}. Instead, it introduces a different angle for tackling content-aware diffusion problems. This new angle allows for unsupervised, non-parametric and online inference.

\section{Experiments}
\subsection{Data and experimental setup}
All data, codes and results are available in open access \footnote{\url{https://github.com/GaelPouxMedard/HOUsToN}}. We consider 3 different network types of 500 nodes each: power-law (\textbf{PL}) \cite{Barabasi2002PLNet}, random Erdös-Renye (\textbf{ER}) \cite{Erdos1960ERgraph} and a real network of hyperlinks between political blogs (\textbf{Blogs}) \cite{Adamic2005BlogsPolitique}. From each network, we sample 5 subnetworks of 250 nodes and assign random weights $\alpha$ between 0 and 1 to their edges. Each of the generated subnetworks is used to propagate one given cluster of information. We then simulate infection cascades on each subnetwork according to the exponential NetRate model. Finally, we associate 5 words drawn from a vocabulary of size 100 to each so-generated event according to its associated subnetwork (or cluster).
We generate a total of 55,000 events $\{ (u_i^c,t_{i}^c,v_{i}^c) \}_{i,c}$ for each network. 

Our hyperparameters are $\theta_0=0.1$ and $\lambda_0^{(k)}=0.001 \ \forall k$. The SMC algorithm considers 4 parallel runs. We consider a constant hazard rate $H(t_i \vert t_j, \alpha_{j,i}) = \alpha_{j,i}$, so the probability of a new event \textit{not} happening decays exponentially with time.

\begin{table}
    \caption{Results on clusters (NMI, ARI) and edges (AUC-ROC, F1, MAE) retrieval. \label{tabMetrics}}
    \centering
    \setlength{\lgCase}{2.3cm}
    \begin{tabular}{|p{0.2\lgCase}|p{0.8\lgCase}|p{0.8\lgCase}|p{0.8\lgCase}|p{0.8\lgCase}|p{0.8\lgCase}|p{0\lgCase}}
    \cline{3-6}
      \multicolumn{0}{c}{\rotatebox[origin=c]{90}{}} &  & \centering Houston & \centering NRxDM & \centering DHP & \centering NetRate & \\
      
     \cline{1-6}
        \centering\multirow{5}{*}{\centering\rotatebox[origin=c]{90}{PL}} & \centering NMI & \centering \textbf{0.809} & \centering 0.669 & \centering 0.449 & \centering - & \\
        
         & \centering ARI & \centering \textbf{0.688} & \centering 0.330 & \centering 0.063 & \centering - & \\
    
     \cdashline{2-6}
        & \centering AUC-ROC & \centering \textbf{0.807} & \centering 0.719 & \centering - & \centering 0.731 & \\
        
         & \centering F1 & \centering \textbf{0.199} & \centering 0.106 & \centering - & \centering 0.005 & \\
         
         & \centering MAE & \centering \textbf{0.267} & \centering 0.338 & \centering - & \centering 0.460 & \\
    
     \cline{1-6}
        \centering\multirow{5}{*}{\centering\rotatebox[origin=c]{90}{ER}} & \centering NMI & \centering \textbf{0.787} & \centering 0.711 & \centering 0.638 & \centering - & \\
        
         & \centering ARI & \centering \textbf{0.631} & \centering 0.488 & \centering 0.411 & \centering - & \\
         
     \cdashline{2-6}
        & \centering AUC-ROC & \centering \textbf{0.849} & \centering 0.800 & \centering - & \centering 0.659 & \\
        
         & \centering F1 & \centering \textbf{0.263} & \centering 0.176 & \centering - & \centering 0.005 & \\
         
         & \centering MAE & \centering \textbf{0.229} & \centering 0.278 & \centering - & \centering 0.481 & \\
    
     \cline{1-6}
        \centering\multirow{5}{*}{\centering\rotatebox[origin=c]{90}{Blogs}} & \centering NMI & \centering \textbf{0.750} & \centering 0.668 & \centering 0.372 & \centering - & \\
        
         & \centering ARI & \centering \textbf{0.609} & \centering 0.365 & \centering 0.023 & \centering - & \\
    
     \cdashline{2-6}
        & \centering AUC-ROC & \centering 0.701 & \centering 0.613 & \centering - & \centering \textbf{0.710} & \\
        
         & \centering F1 & \centering \textbf{0.168} & \centering 0.087 & \centering - & \centering 0.005 & \\
         
         & \centering MAE & \centering \textbf{0.374} & \centering 0.444 & \centering - & \centering 0.499 & \\
     \cline{1-6}
    \end{tabular}
\end{table}
\subsection{Results}
We compare to 3 similar baselines used as ablation tests: \textbf{Dirichlet-Hawkes process (DHP)} \cite{Du2015DHP} clusters textual data by using temporal dynamics, and does not consider structure; \textbf{NetRate} \cite{GomezRodriguez2011NetRate} infers a dynamic network based on observed cascades without considering their content; \textbf{NetRate x Dirichlet-Multinomial (NRxDM)} first uses textual information to infer clusters, and only then infers the underlying subnetwork for each cluster, in the same fashion as \cite{Du2013TopicCascade,He2015HawkesTopic}.
When applicable, we evaluate on a classification task (scores NMI and ARI with respect to the clusters used for data generation) and a network inference task (AUC-ROC, F1 and MAE on the true edges, same metrics as in \cite{GomezRodriguez2011NetRate}).

We see in Table~\ref{tabMetrics} that Houston consistently outperforms methods that do not consider jointly text, time and structure of the network. 
To summarize, NRxDM only considers textual information to build clusters, making the network inference miss a great deal of temporal and structural information. DHP considers textual information and temporal dynamics, but misses the structural information. NetRate does not consider textual data and infers the network based on temporal dynamics only. Houston bridges the gap between these models, by making a joint use of textual, temporal and structural information.

As an illustration of what Dirichlet-Processes can yield on real-world data, we draft its application to the Memetracker dataset \cite{Leskovec2009Memetracker} in Fig.~\ref{fig:task-houston} (bottom). We retrieve the diffusion network associated to meme clusters and observe diverse spreading dynamics. Topics spread in distinct parts of the global network, and mostly do so through a reduced set of densely connected nodes, as shown in \cite{GomezRodriguez2013InfoPath}.

\section{Conclusion}
In this paper, we propose the Dirichlet-Survival process as an alternative way to jointly model textual, temporal and structural information in spreading processes. Ablation tests demonstrate the relevance of the proposed approach. 
As a prior, the Dirichlet-Survival process can add a dynamic network dimension to any sequential Bayesian model; it could be coupled to models that account for any type of clustering (e.g. images, time series, labels), or simply more refined language models. Its introduction opens new perspectives on traditional machine learning problems, including topic-dependent spreading processes on networks.

%
%
%
\bibliographystyle{splncs04}
\bibliography{Bibliography}
\end{document}